\documentclass[conference,letterpaper]{IEEEtran}

\usepackage{hyperref}
\usepackage[utf8]{inputenc} 
\usepackage[T1]{fontenc}
\usepackage{url}
\usepackage{ifthen}

\usepackage{cite}
\usepackage[cmex10]{amsmath}

\usepackage{booktabs}       % professional-quality tables
\usepackage{amsfonts}       % blackboard math symbols
\usepackage{nicefrac}       % compact symbols for 1/2, etc.
\usepackage{microtype}      % microtypography
\usepackage{amsmath,amssymb,amsfonts}
\usepackage{algorithm}
\usepackage{algpseudocode}
\usepackage{xspace}
\usepackage{xcolor}         % colors

\usepackage{enumitem}
\usepackage{bm}
\usepackage{amsmath}
\usepackage{amsthm}
\usepackage{caption}
\usepackage{amssymb}
\usepackage{subcaption}
\usepackage{wrapfig}
\usepackage{import}
\usepackage{tcolorbox}

\def\enc{\mathbf{u}_\textrm{enc}}
\def\dec{\mathbf{u}_\textrm{dec}}

\def\func{\mathbf{f}}
\def\stset{\mathcal{F}}

\newtcolorbox{mathbox}[1][]{colback=gray!20, #1}

\tcbset{
  width=(\linewidth-4mm),
  center,                 
  before skip=.7em,     
  after skip=.7em,       
  colframe=black!75!black,
  colback=black!5!gray
}

\begin{document}
\title{ITW 2026 Paper Template}

\title{Manifold-Aware General Coded Computing for Straggler-Resilient Distributed Computing} 

\author{%
  \IEEEauthorblockN{Parsa Moradi}
  \IEEEauthorblockA{University of Minnesota, Twin Cities\\
                    Minneapolis, MN, USA\\
                    moradi@umn.edu}
  \and
  \IEEEauthorblockN{Mohammad Ali Maddah-Ali}
  \IEEEauthorblockA{University of Minnesota, Twin Cities\\
                    Minneapolis, MN, USA\\
                    maddah@umn.edu}
}
\maketitle

\begin{abstract}
% General coded computing (GCC) provides a learning-theoretic framework for the straggler-resilient distributed evaluation of general high-dimensional nonlinear functions. In GCC, the master node constructs an encoder curve to an input batch, samples this curve to generate coded inputs, assigns each coded input to a worker, and constructs a decoder curve to the outputs returned by the non-straggling workers to recover the desired function evaluations. 
Existing coded-computing designs do not explicitly exploit the intrinsic structure of the input data. In communication systems, statistical structure and redundancy are often removed through source coding (or compression) before channel coding is applied. This principle, however, does not transfer directly to coded computation. In many computational tasks, particularly in machine learning, the structure of the data is precisely what the computation seeks to exploit to infer outputs or learn meaningful patterns. Consequently, coded-computing schemes should preserve and leverage this structure in their code design, rather than ignoring or eliminating it through source coding.

This observation motivates a different perspective on code construction. In many channel-coding schemes, such as Reed-Solomon codes, coded symbols are generated by evaluating a low-dimensional algebraic representation at selected points. In contrast, many high-dimensional datasets naturally concentrate near low-dimensional manifolds. In this paper, we exploit this intrinsic geometry by designing coded samples that follow the natural manifold of the data, rather than imposing an artificial low-dimensional structure unrelated to the data distribution. Inspired by graph-based manifold learning, we propose a manifold-aware encoding strategy for general coded computing (GCC). Experiments on neural network inference and high-dimensional polynomial evaluation demonstrate that the proposed strategy consistently and significantly reduces the mean squared recovery error under straggling compared with standard GCC.

% Inspired by graph-based manifold learning, we propose a manifold-aware encoding strategy for GCC. The proposed method constructs a distance-weighted graph over the input batch and orders its vertices along a short Hamiltonian path. This path determines a geometry-aware assignment of the input samples to the ordered encoder design points, encouraging the fitted encoder curve, and hence the coded inputs sampled from it, to remain close to the underlying data manifold. We further show that the Hamiltonian-path length controls an explicit upper bound on the end-to-end GCC loss, providing a theoretical justification for the proposed ordering strategy. Experiments on neural network inference and high-dimensional polynomial evaluation demonstrate that the proposed strategy consistently reduces the mean-squared recovery error under straggling compared with standard GCC.
\end{abstract}

\section{Introduction}

Distributed computing is a fundamental approach to accelerate large-scale computational workloads, including machine learning training and inference. In a typical master-worker architecture, a master node seeks to evaluate a function on a batch of input data by distributing computational tasks across multiple worker nodes. In practice, however, the overall completion time is often dominated by \emph{stragglers}: workers that are delayed, overloaded, or fail to return their outputs before a prescribed deadline. Developing effective mechanisms for mitigating the impact of stragglers is therefore a central problem in distributed computing.

Coded computing addresses this challenge by introducing structured redundancy into tasks assigned to workers~\cite{jahani2022berrut,yu2020straggler,karakus2017straggler,soleymani2022approxifer,yu2019lagrange,moradicoded,moradi2025generalprob,Abraham1984,speed,high}. Rather than assigning the original inputs directly, the master node generates a collection of \emph{coded inputs} via a designed encoding procedure. Each worker evaluates the prescribed target function on its coded input and returns the resulting \emph{coded output}. The master node then applies a decoding procedure to recover the desired function evaluations from the outputs returned by the non-straggling workers.

Classical coded-computing schemes draw heavily on algebraic coding theory and provide rigorous straggler-resilience guarantees for structured computations, such as matrix multiplication and polynomial evaluation~\cite{yu2019lagrange,fahim2019numerically,fahimappx,yu2020straggler,yu2017polynomial,yu2020entangled,opt-recovery,short}. Many modern computational workloads, however, involve general high-dimensional nonlinear functions, including deep neural networks, that lack the algebraic structure these schemes require. This limitation has motivated the development of more flexible coded-computing approaches for general nonlinear functions~\cite{jahani2022berrut,soleymani2022approxifer,SoAppxMPC,codedpri}. Among these approaches, \emph{General Coded Computing} (GCC) provides a learning-theoretic framework for designing straggler-resilient coded-computing schemes for arbitrary functions~\cite{moradicoded,moradi2025generalprob}. GCC formulates coded computation as an end-to-end approximation problem and designs the encoder and decoder by optimizing a loss function that directly measures the discrepancy between the desired function evaluations and the estimates recovered from the non-straggling worker outputs.

Nevertheless, existing coded-computing approaches generally overlook an important property of the input data: its intrinsic statistical structure. This structure often implies substantial redundancy in the data. In communication systems, this redundancy is typically removed first through source coding (or compression) before controlled redundancy is introduced via channel coding to enable reliable communication over an uncertain or noisy channel~\cite{shannon1948mathematical,cover2006elements}. This classical separation is supported by Shannon's source-channel separation theorem, which establishes, under standard assumptions for point-to-point memoryless channels, that separate source and channel coding can achieve optimal performance asymptotically~\cite{cover2006elements}.

However, this separation principle does not carry over directly to coded computing. Consider machine learning computations, which have been a major motivation for the coded-computing paradigm. Many such tasks are specifically designed to exploit the statistical and geometric structure of the input data to infer outputs or learn meaningful patterns. Applying source coding solely to eliminate intrinsic data redundancy before designing the coded-computing scheme may therefore discard or obscure precisely the structure the computation aims to exploit. Moreover, it is not evident that a strict separation between source compression and coding for reliable computation is desirable in this setting. Instead, these observations suggest a more integrated design principle:

\begin{mathbox}
\emph{Coded-computing schemes should explicitly embrace and exploit the intrinsic structure of the data, rather than ignoring or treating it merely as redundancy to be removed prior to coding.}
\end{mathbox}

\begin{figure}[t]
    \centering
    \includegraphics[width=1.0\linewidth]{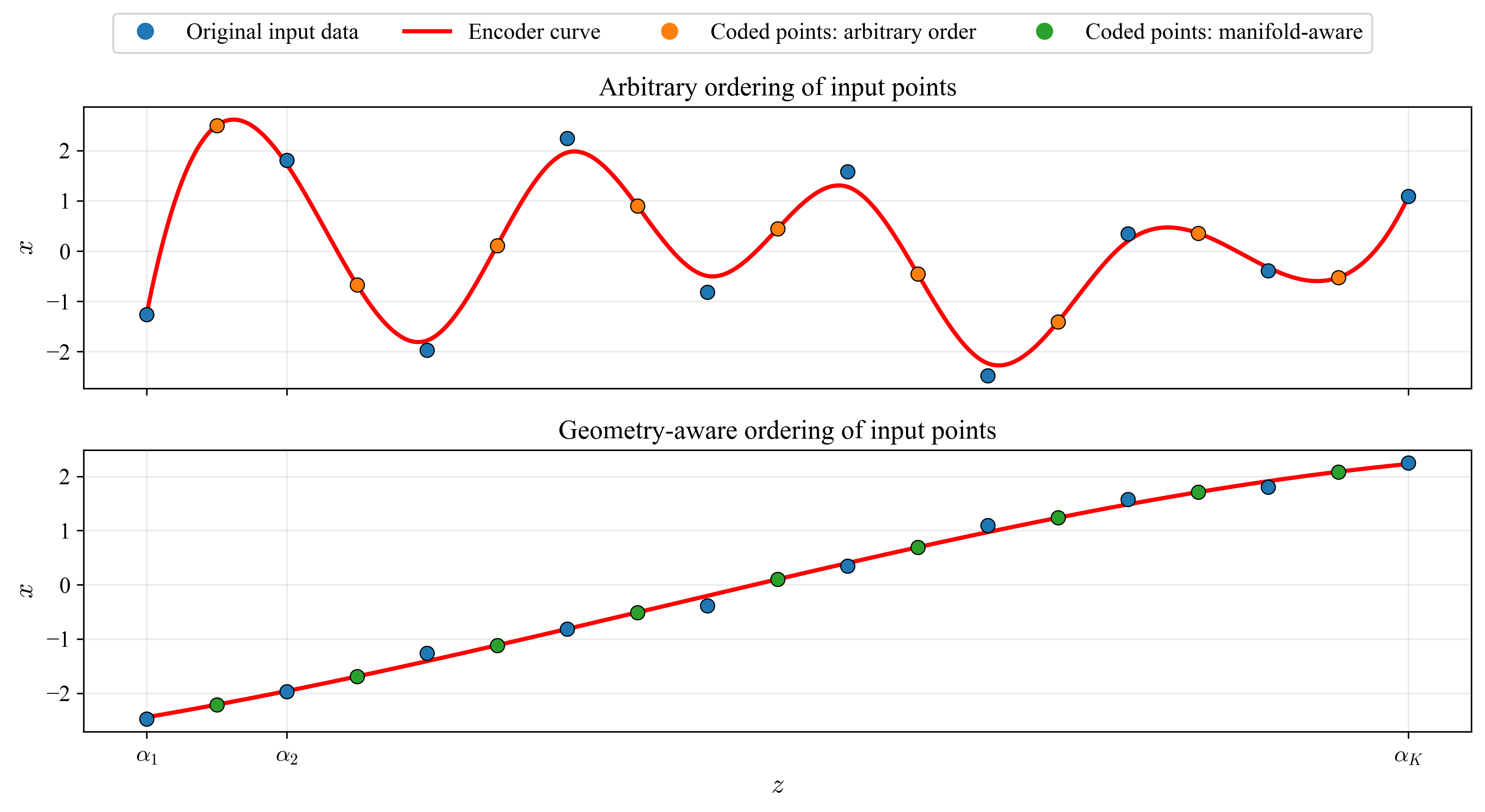}
    \caption{Illustration of the importance of exploiting data structure in coded computing. Top: standard GCC. Bottom: GCC with a structure-aware encoder that exploits the intrinsic geometry of the data.}
\label{fig:order_concept2}
\vspace{-10pt}
\end{figure}

Let us illustrate this consideration through a simple toy example. We first provide a high-level overview of how GCC operates. GCC selects a set of fixed, ordered scalar locations on a one-dimensional interval, referred to as \emph{encoder design points}, and associates the input samples with these locations (see the blue points in Fig.~\ref{fig:order_concept2}, top). The master node then constructs an \emph{encoder curve} that maps each encoder design point approximately to its associated input sample (red curve) and samples this curve at a second set of fixed locations, one for each worker, to generate coded inputs (green points). After the workers evaluate the target function on these coded inputs, the master node constructs a \emph{decoder curve} from the returned coded outputs and evaluates it at the encoder design points to recover the desired function values.

% This curve-based viewpoint exposes an important design question that has not been explicitly addressed in existing GCC constructions. The encoder design points have an intrinsic order along the one-dimensional interval, whereas the input batch is generally an unordered collection of points in a high-dimensional space. Consequently, constructing the encoder requires choosing an assignment between the input samples and the ordered encoder design points. Existing GCC implementations typically inherit this assignment from the original indexing of the input batch rather than optimizing it according to the geometry of the data. However.

Let us now design the encoder by explicitly exploiting the structure of the data. We first order the data points and then assign them to the \emph{encoder design points} according to this ordering; see Fig.~\ref{fig:order_concept2} (bottom). As illustrated in the figure, the resulting encoding curve, shown in red, is substantially smoother. The coded symbols, shown in green, are again obtained by sampling this curve. Heuristically, a smoother encoding curve can be represented more accurately from a limited number of sampled points. Equivalently, fewer coded symbols may be sufficient to reconstruct the red curve, and hence to recover the original data points more accurately. Indeed, the theory of generalized coded computing suggests that a smoother encoding function leads to more accurate coded-computing results~\cite{moradicoded,moradi2025generalprob}.

This toy example highlights the importance of exploiting the intrinsic structure of the data in code design. Many real-world high-dimensional datasets exhibit precisely the structure that can be leveraged for this purpose. According to the manifold hypothesis, such datasets often concentrate near a low-dimensional manifold embedded in the ambient high-dimensional space~\cite{fefferman2016testing}. This low-dimensional geometry can be viewed as a form of intrinsic redundancy.

\begin{figure*}[t]
    \centering
    \includegraphics[width=1.0\linewidth]{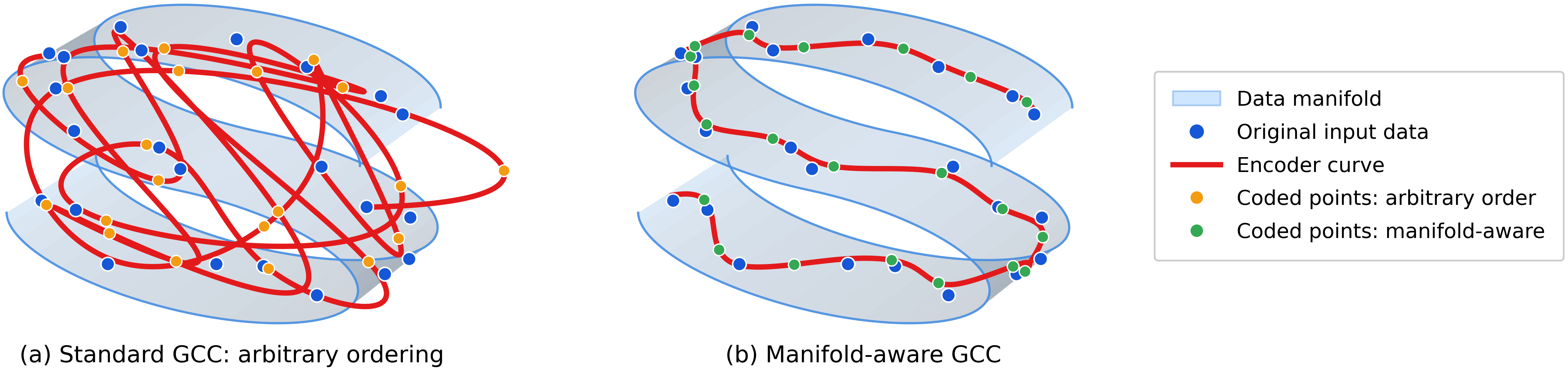}
    \caption{Illustration of standard and manifold-aware GCC encoding. Original input samples (blue) lie on a low-dimensional data manifold. With an arbitrary ordering, the fitted encoder curve (red) makes large transitions through the ambient space, causing many coded inputs (orange) to lie away from the data manifold. In contrast, the manifold-aware ordering encourages the encoder curve to follow the data geometry, keeping the resulting coded inputs (green) in the vicinity of the manifold.}
    \label{fig:order_concept}
\end{figure*}

Fig.~\ref{fig:order_concept} illustrates an example of such a manifold, where the input data points lie in its vicinity. Fig.~\ref{fig:order_concept}(a) shows a code design that does not exploit the intrinsic structure of the data. As a result, the coded data symbols are sampled from a relatively complex encoding curve, which can lead to a higher approximation error in the decoding process. In contrast, Fig.~\ref{fig:order_concept}(b) illustrates a code design that exploits this structure and achieves a more accurate approximation.

As shown in Fig.~\ref{fig:order_concept}(b), when coded computing respects the structure of the data, the coded symbols remain in the vicinity of the data manifold. This observation highlights a key perspective: because the target function is evaluated on coded symbols rather than directly on original data, each coded representation generated by the encoder must constitute a meaningful input to the target function. If the coded-computing scheme fails to respect the data's intrinsic structure, the resulting coded symbols may fall outside the regions of the input space where the target function operates reliably.

For example, in an inference task where the target function is a machine learning model trained on data drawn from a specific domain distribution, coded symbols that stay near the data manifold are more likely to lie in data-supported regions where the model performs well. In contrast, coded symbols that deviate substantially from this manifold may act as out-of-distribution inputs, leading to unreliable predictions after decoding. Similarly, when computing the gradient of a loss function for training, generating coded samples close to the manifold helps ensure training occurs on inputs representative of the underlying data domain.

% Manifold learning is based on the premise that high-dimensional data often possess a lower-dimensional intrinsic structure. In particular, although each data point is represented in a high-dimensional ambient space, the collection of data points typically concentrates near a smooth low-dimensional manifold.

Uncovering and learning the low-dimensional intrinsic structure near which data points concentrate in a high-dimensional ambient space is referred to as \emph{manifold learning}. Classical examples include Isomap~\cite{tenenbaum2000global} and Laplacian Eigenmaps~\cite{belkin2003laplacian}.
% This perspective suggests that, rather than treating the input samples as an unstructured collection of points in the ambient space, one can design computational and coding schemes that explicitly exploit their intrinsic geometric organization.  
A common approach in manifold learning is to represent relationships among high-dimensional samples using a graph whose nodes correspond to data points and whose edge weights are determined by pairwise distances~\cite{tenenbaum2000global,belkin2003laplacian,mcinnes2018umap,faigenbaum2021manifold}. Nearby samples are interpreted as locally related, and the resulting graph provides a discrete proxy for the geometry of the underlying data manifold.

Motivated by these observations, this paper proposes a \emph{manifold-aware GCC} encoding strategy. Given an input batch, the proposed method constructs a distance-weighted graph whose vertices correspond to input samples and whose edge weights reflect pairwise distances. It then uses a short Hamiltonian path to obtain a one-dimensional traversal of the empirical data geometry and assigns the samples to ordered encoder design points accordingly. The GCC encoder is subsequently constructed using this geometry-aware assignment. As a result, consecutive portions of the encoder curve connect nearby samples, encouraging coded inputs sampled from the curve to remain closer to the underlying data manifold. Thus, the proposed strategy introduces input ordering as an additional code-design variable while preserving the standard GCC pipeline.

We evaluate the proposed method on two representative classes of high-dimensional nonlinear computations: neural network inference using LeNet5~\cite{lecun1998gradient} and high-dimensional polynomial evaluation. Experimental results show that the proposed manifold-aware ordering consistently improves GCC recovery performance under straggling, yielding lower mean squared recovery error than standard GCC.

\section{Problem Formulation and GCC Background}
\label{sec:prelim}

We consider a distributed computing system consisting of one master node and \(N\) worker nodes. The master node seeks to compute a batch of function evaluations
\(
    \{\func(\mathbf{x}_k)\}_{k=1}^K,
\)
where each input data point satisfies \(\mathbf{x}_k\in\mathbb{R}^d\), and the target function is
\(
    \func:\mathbb{R}^d\to\mathbb{R}^m.
\)
The function \(\func(\cdot)\) may represent a general high-dimensional computation, ranging from a simple vector-valued function to a deep neural network used for inference.

A central challenge in this setting is the presence of \emph{straggling} workers. Due to delays, congestion, or failures, some workers may not return their assigned computations before a prescribed deadline. Let
\(
    \mathcal{F}\subseteq [N] :=\{1, \dots,N\}
\)
denote the set of non-straggling workers whose outputs are received by the master node on time. The set \(\mathcal{F}\) may vary across computation rounds, and we model it as a random subset drawn from a distribution \(P_{\mathcal{F}}\) over subsets of \([N]\). We next describe the standard GCC framework. 

\subsection{Standard General Coded Computing}

Standard GCC consists of three stages: (i) encoding at the master node, (ii) computation at the worker nodes, and (iii) decoding at the master node.

\paragraph{Encoding}
The master node selects \(K\) ordered encoder design points
\(
    \alpha_1 < \alpha_2 < \cdots < \alpha_K
\)
in a bounded interval \(\Omega\subset\mathbb{R}\). It then constructs a second-order smoothing spline~\cite{wahba1975smoothing,wahba1990spline}
\(\enc:\Omega\to\mathbb{R}^d\) using the data-design pairs
\(
    \{(\alpha_k,\mathbf{x}_k)\}_{k=1}^K
\)
such that
\(
    \enc(\alpha_k)\approx \mathbf{x}_k
\)
for \(k\in[K]:=\{1,\ldots,K\}\). The master node also selects \(N\) ordered decoder design points
\(
    \beta_1 < \beta_2 < \cdots < \beta_N
\)
in \(\Omega\). For each worker \(n\in[N]\), the master node generates the coded input
\(
    \widetilde{\mathbf{x}}_n = \enc(\beta_n)
\)
and sends \(\widetilde{\mathbf{x}}_n\) to worker \(n\). Because the encoder is constructed using the entire input batch, each coded input \(\widetilde{\mathbf{x}}_n\) generally depends on all original input samples.

\paragraph{Computation}
Each worker applies the target function to its assigned coded input. If worker \(n\) is non-straggling, it returns
\(
    \mathbf{y}_n
    =
    \func(\widetilde{\mathbf{x}}_n)
    =
    \func(\enc(\beta_n))
\)
to the master node before the deadline. Let \(\mathcal{F}\subseteq[N]\) denote the set of non-straggling workers. The master node therefore observes
\(
    \{(\beta_n,\mathbf{y}_n)\}_{n\in\mathcal{F}}
\).

\paragraph{Decoding}
Using the outputs returned by the non-straggling workers, the master node fits another second-order smoothing spline
\(
    \dec:\Omega\to\mathbb{R}^m
\)
to the observed pairs
\(
    \{(\beta_n,\mathbf{y}_n)\}_{n\in\mathcal{F}}
\).
The decoder is intended to approximate target function evaluations along the encoder curve; in particular,
\(
    \dec(\beta_n)
    \approx
    \func(\enc(\beta_n))
\)
for \(n\in\mathcal{F}\). The master node then estimates the desired function evaluations by evaluating the decoder at the encoder design points:
\(
    \widehat{\func}_{\mathcal{F}}(\mathbf{x}_k)
    :=
    \dec(\alpha_k)
\)
for \(k\in[K]\). When both the encoder and decoder are accurate, one expects
\(
    \dec(\alpha_k)
    \approx
    \func(\enc(\alpha_k))
    \approx
    \func(\mathbf{x}_k)
\).

The performance of the scheme is measured by the mean squared recovery error averaged over the straggling pattern:
\begin{align}
    \mathcal{L}
    :=
    \mathbb{E}_{\mathcal{F}\sim P_{\mathcal{F}}}
    \left[
    \frac{1}{K}
    \sum_{k=1}^K
    \left\|
    \widehat{\func}_{\mathcal{F}}(\mathbf{x}_k)
    -
    \func(\mathbf{x}_k)
    \right\|_2^2
    \right].
    \label{eq:main_loss}
\end{align}
This objective directly captures the goal of coded computing: the master node should accurately recover the function values using only the outputs returned by non-straggling workers.

The use of second-order smoothing splines in standard GCC arises from a principled, optimization-based design. Specifically, the authors of~\cite{moradicoded} optimize a tractable upper bound on~\eqref{eq:main_loss} and show that, when chosen from second-order Sobolev spaces---function spaces with square-integrable derivatives up to order two---the optimal encoder and decoder are second-order smoothing splines fitted to the input data and returned worker outputs, respectively.

\begin{algorithm}[t]
\vspace{6pt}
\caption{Manifold-Aware GCC}
\label{alg:manifold_gcc}
\begin{algorithmic}[1]
\Statex \textbf{Input:} Input batch $\{\mathbf{x}_k\}_{k=1}^K$, encoder design points $\{\alpha_k\}_{k=1}^K$, decoder design points $\{\beta_n\}_{n=1}^N$
\Statex \textbf{Output:} Coded inputs $\{\widetilde{\mathbf{x}}_n\}_{n=1}^N$ and recovered estimates $\{\widehat{\func}_{\mathcal{F}}(\mathbf{x}_k)\}_{k=1}^K$
\State Construct graph weights $w_{ij}=\|\mathbf{x}_i-\mathbf{x}_j\|_2$ for $i,j\in[K]$.
\State Let \(S_K\) denote the set of all permutations of \([K]\), and find a short Hamiltonian path
\[
    \pi
    \approx
    \arg\min_{\sigma\in S_K}
    \sum_{j=1}^{K-1}
    w_{\sigma(j),\sigma(j+1)} .
\]
\State Fit the encoder smoothing spline $\enc:\Omega\to\mathbb{R}^d$ using the ordered pairs
\(
    \{(\alpha_j,\mathbf{x}_{\pi(j)})\}_{j\in[K]} .
\)
\For{$n=1,\ldots,N$}
    \State Set $\widetilde{\mathbf{x}}_n=\enc(\beta_n)$ and send it to worker $n$.
\EndFor
\State Receive $\{\func(\widetilde{\mathbf{x}}_v)\}_{v\in\mathcal{F}}$ from the non-straggling workers.
\State Fit the decoder smoothing spline $\dec:\Omega\to\mathbb{R}^m$ using the returned pairs
\(
    \{(\beta_v,\func(\widetilde{\mathbf{x}}_v))\}_{v\in\mathcal{F}} .
\)
\State Return $\widehat{\func}_{\mathcal{F}}(\mathbf{x}_{\pi(j)})=\dec(\alpha_j)$ for all $j\in[K]$.
\end{algorithmic}
\end{algorithm}

\section{Manifold-Aware General Coded Computing}
In standard GCC, coded inputs are generated by sampling an encoder curve \(\enc(z)\) at \(\{\beta_j\}_{j\in[N]}\), where they are evaluated by the target function (\(\{\func(\enc(\beta_j))\}_{j\in[N]}\)). Ideally, the encoder curve \(\enc(z)\) should be as smooth as possible, as a smoother encoder generally leads to a smoother composition \(\func(\enc(z))\), which in turn reduces approximation error during decoding~\cite{moradicoded}.

Conversely, when input data concentrate near a low-dimensional manifold, coded inputs should ideally lie on or near this manifold. This ensures that the coded inputs remain in the vicinity of the input data manifold where the target function is intended to operate. In contrast, coded inputs far from the data manifold may be out-of-distribution, rendering their function evaluations less informative for recovering desired outputs.

This observation highlights an important design choice in GCC. The encoder design points \(\alpha_1 < \cdots < \alpha_K\) are ordered along a one-dimensional interval, whereas the input batch \(\{\mathbf{x}_k\}_{k=1}^K\) is generally unordered. To obtain a smoother encoding function that remains close to the data manifold, input samples close to one another should be assigned to nearby encoder design points. If consecutive design points are assigned to distant input samples, the resulting encoder curve must make large transitions through ambient space, potentially passing through regions far from the underlying data manifold. 
% In contrast, assigning consecutive design points to nearby samples encourages the encoder curve to follow the local geometry of the data.

% In standard GCC, the encoder design points
% \(\alpha_1<\cdots<\alpha_K\) are ordered, while the input batch
% \(\{\mathbf{x}_k\}_{k=1}^K\) is generally unordered. Therefore, fitting the encoder implicitly requires choosing an assignment between data points and design points. Equivalently, for a permutation \(\pi\) of \([K]\), the encoder is fitted to the ordered pairs
% \(
%     \{(\alpha_j,\mathbf{x}_{\pi(j)})\}_{j=1}^K .
% \)
% This assignment is not a minor implementation detail: it directly determines the geometry and complexity of the encoder curve.

Figs.~\ref{fig:order_concept2} and~\ref{fig:order_concept} illustrate this effect in a one-dimensional example. The same set of input points can induce a highly oscillatory spline when assigned to the encoder design points in an arbitrary order. In contrast, ordering points according to their geometry produces a much simpler, nearly linear encoder curve. More generally, assigning consecutive design points \(\alpha_j\) and \(\alpha_{j+1}\) to distant data points forces the encoder curve to bridge a large gap in the empirical data geometry. Coded inputs sampled along such transitions may pass through regions with little data support, increasing the variation of the encoder and the induced computation \(\func\circ\enc\). This can lead to higher decoding error from non-straggling worker outputs.

\begin{figure}[t]
    \centering
    \includegraphics[width=1.0\linewidth]{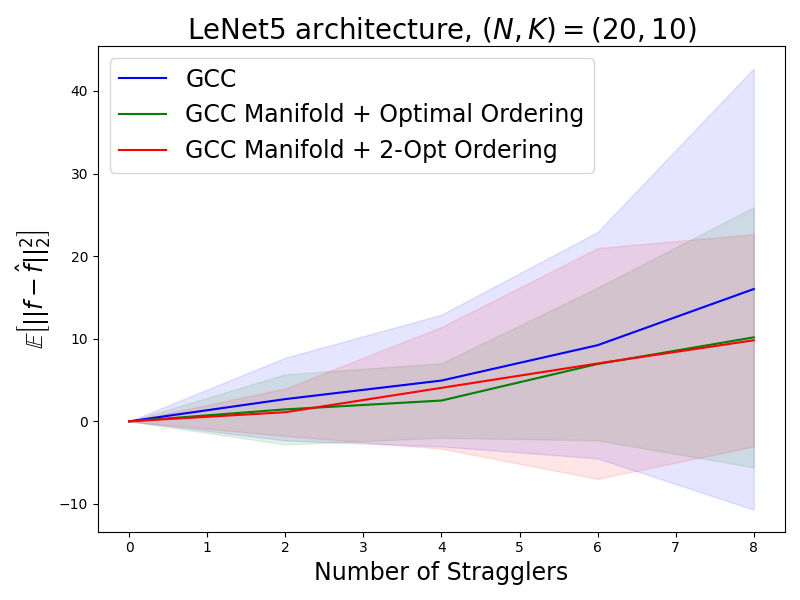}
    \caption{Comparison of standard GCC, manifold-aware GCC with exact shortest-Hamiltonian-path ordering, and manifold-aware GCC with 2-opt ordering for LeNet5 with \((N,K)=(20,10)\). The plot reports the mean squared recovery error as a function of the number of stragglers. The 2-opt ordering closely tracks the performance of the exact optimal ordering, indicating that the heuristic achieves comparable recovery performance while being substantially more efficient to compute. Shaded regions indicate \(95\%\) confidence intervals over repeated trials.}
    \label{fig:ham_path_comp}
\end{figure}

\begin{figure*}[t]
    \centering
    \begin{subfigure}[t]{0.23\textwidth}
        \centering
        \includegraphics[width=\textwidth]{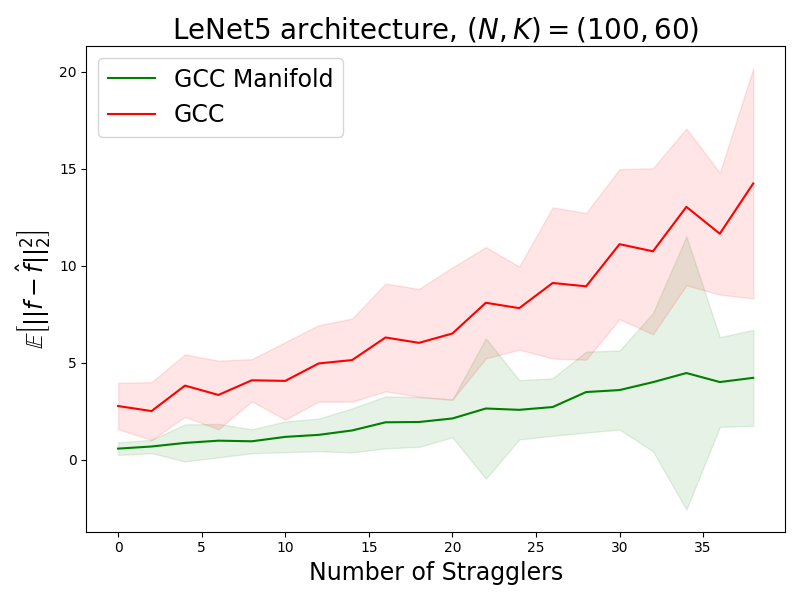}
        \caption{Mean squared recovery error for LeNet5.}
        \label{fig:lenet5_mse}
    \end{subfigure}
    \hfill
    \begin{subfigure}[t]{0.23\textwidth}
        \centering
        \includegraphics[width=\textwidth]{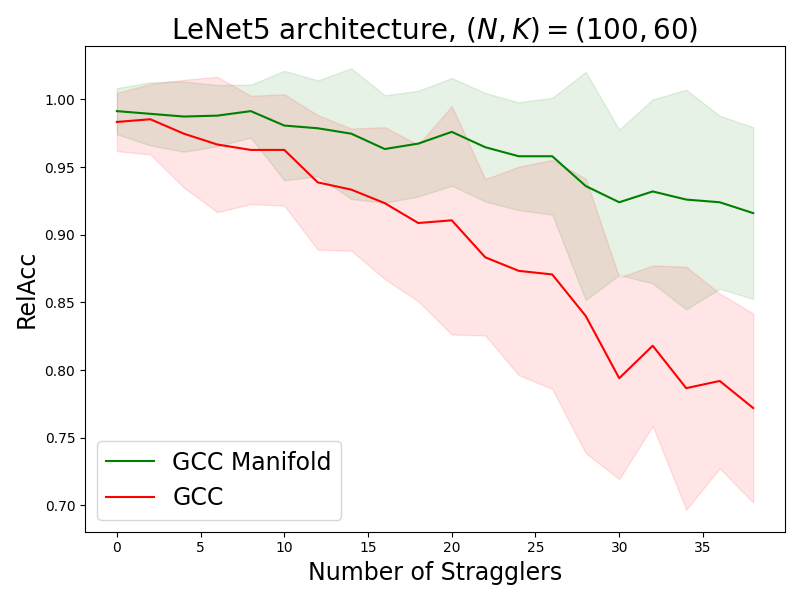}
        \caption{Relative accuracy for LeNet5.}
        \label{fig:lenet5_relacc}
    \end{subfigure}
    \hfill
    \begin{subfigure}[t]{0.23\textwidth}
        \centering
        \includegraphics[width=\textwidth]{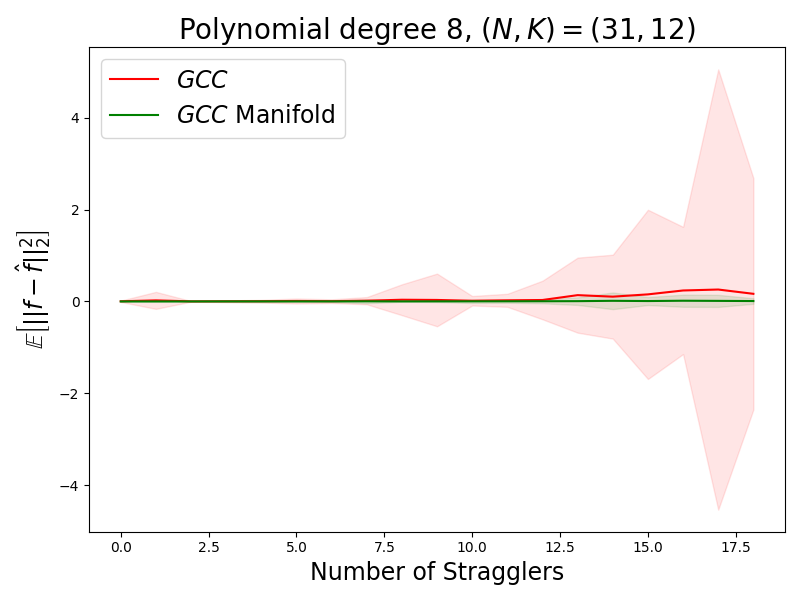}
        \caption{Mean squared recovery error for the degree-\(8\) polynomial task.}
        \label{fig:poly_mse}
    \end{subfigure}
    \hfill
    \begin{subfigure}[t]{0.23\textwidth}
        \centering
        \includegraphics[width=\textwidth]{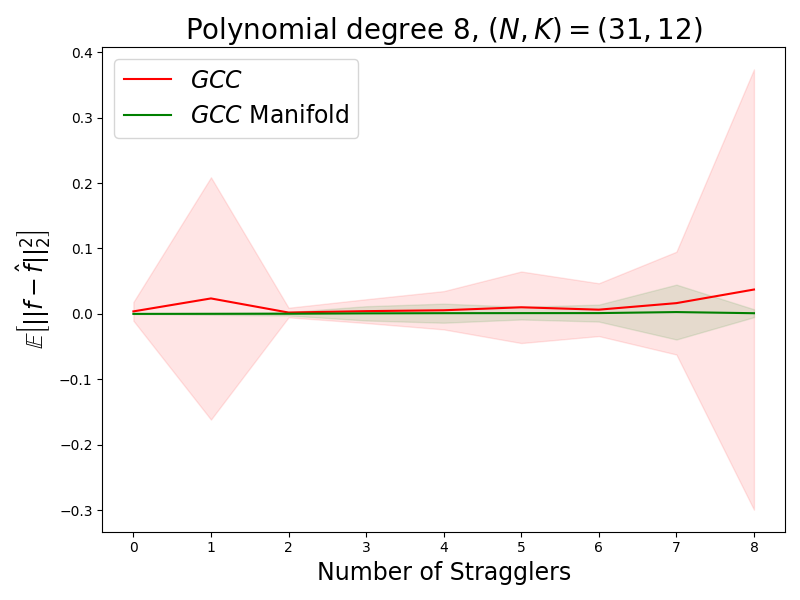}
        \caption{Zoomed view of the polynomial recovery error.}
        \label{fig:poly_mse_zoom}
    \end{subfigure}

    \caption{Performance comparison between standard GCC and manifold-aware GCC as the number of stragglers increases. The first two panels report results for LeNet5 with \((N,K)=(100,60)\), where we evaluate both mean squared recovery error and relative accuracy. The last two panels report results for a degree-\(8\) polynomial task with \((N,K)=(31,12)\), showing the mean squared recovery error and a zoomed view. Shaded regions indicate \(95\%\) confidence intervals over repeated trials.}
    \label{fig:gcc_comparison}
\end{figure*}

Motivated by this principle, we treat the assignment as an additional code-design variable. Equivalently, for a permutation \(\pi\) of \([K]\), the encoder is fitted using the ordered pairs
\(
    \{(\alpha_j,\mathbf{x}_{\pi(j)})\}_{j=1}^K
\).
Our objective is to select \(\pi\) so that this ordering reflects the empirical geometry of the input data. Given the data points \(\{\mathbf{x}_k\}_{k=1}^K\), we construct a weighted graph
\(
    \mathcal{G} = ([K], \mathcal{E}, w),
\)
where each node corresponds to an input data point and the edge weight between nodes \(i\) and \(j\) is given by
\(
    w_{ij} = \|\mathbf{x}_i-\mathbf{x}_j\|_2.
\)
More generally, \(w_{ij}\) may be chosen as a distance in a feature space. We then seek a permutation \(\pi\) of \([K]\) that minimizes total path length:
\begin{align}\label{eq:ham_path}
    \pi^* = \arg\min_{\pi} 
    \sum_{j=1}^{K-1}
    w_{\pi(j),\pi(j+1)}.
\end{align}
Finding the exact minimizer of~\eqref{eq:ham_path} is the shortest Hamiltonian path problem, a path variant of the traveling salesman problem (TSP). Exact dynamic programming methods, such as the Bellman--Held--Karp algorithm, solve TSP-type problems in \(O(K^2 2^K)\) time and \(O(K 2^K)\) memory~\cite{bellman1962dynamic}. Therefore, exact optimization is impractical as a preprocessing step for GCC when \(K\) is moderately large.

% In practice, GCC does not require the globally shortest path; it only requires an ordering that avoids the long jumps produced by a random permutation. For this reason, we use a local-search heuristic based on 2-opt moves~\cite{croes1958method}. A 2-opt step replaces two edges in the current route whenever the exchange decreases the total path length. One full scan of the 2-opt neighborhood has \(O(K^2)\) cost after the pairwise distance matrix has been computed, and the method is repeated until no improving 2-opt move is found. Thus, the overall cost is \(O(K^2d + T K^2)\), where \(O(K^2d)\) is the cost of computing pairwise distances in \(\mathbb{R}^d\), and \(T\) is the number of local-search passes. Therefore, since the encoder and decoder of GCC are efficiently computed with complexities of $\mathcal{O}((N+k)d)$ and $\mathcal{O}((|\stset|+k)m)$, respectively, the total computational complexiyt of the the proposed approache is $\mathcal{O}(K^2d + TK^2 + Nd + (N+K)m)$.

In practice, manifold-aware GCC does not require a globally optimal Hamiltonian path; it only requires an ordering that avoids the long transitions induced by an arbitrary permutation (see Fig.~\ref{fig:ham_path_comp}). We therefore employ a local-search heuristic based on 2-opt moves~\cite{croes1958method}. A 2-opt move replaces two edges of the current path whenever the resulting exchange reduces total path length, repeating this process until no further improving move is found. Finally, after computing the ordering \(\pi^*\), the encoder is fitted according to
\(
    \enc(\alpha_j) \approx \mathbf{x}_{\pi^*(j)}
\)
for \(j\in[K]\), while the rest of the GCC pipeline remains unchanged.

\subsection{Computational Complexity}
After the pairwise distance matrix has been computed in  \(\mathcal{O}(K^2d)\), one full scan of the 2-opt neighborhood costs \(\mathcal{O}(K^2)\). Thus, if \(T\) denotes the number of local-search passes, the ordering stage has computational complexity of
\(
    \mathcal{O}\bigl(K^2d + K^2T\bigr).
\)
In addition, the standard GCC encoder and decoder can be computed with complexities \(\mathcal{O}((N+K)d)\) and \(\mathcal{O}((|\stset|+K)m)\), respectively~\cite{moradicoded}. Therefore, the total master-side computational complexity of the proposed manifold-aware GCC scheme is
\(
    \mathcal{O}\bigl(
        K^2d
        +
        TK^2
        +
        (N+K)d
        +
        (|\stset|+K)m
    \bigr).
\)
Since \(|\stset|\leq N\) and \(K^2d\) dominates \(Kd\), this complexity can be upper-bounded as
\(
    \mathcal{O}\bigl(
        K^2d
        +
        TK^2
        +
        Nd
        +
        (N+K)m
    \bigr).
\)

\section{Experimental Results}
\label{sec:experiments}

We evaluate the effectiveness of the proposed manifold-aware GCC (denoted as GCC-Manifold) scheme on two computational tasks. The first task is neural network inference using LeNet5~\cite{lecun1998gradient}, a convolutional neural network with approximately \(6\times 10^4\) trainable parameters for handwritten digit classification~\cite{lecun2010mnist}. The second task is the evaluation of a high-dimensional polynomial function. These two settings allow us to test the proposed ordering strategy on both a practical nonlinear machine learning model and a classic high-dimensional function.

In all experiments, we compare standard GCC with the proposed manifold-aware GCC. For the encoder and decoder design points, we choose \(\{\alpha_k\}_{k=1}^K\) and \(\{\beta_n\}_{n=1}^N\) as Chebyshev points of the first and second kind, respectively, with \(\Omega=(-1,1)\). This choice is motivated by the desirable interpolation properties of Chebyshev grids, which help reduce oscillatory behavior near the boundaries of the interpolation interval \(\Omega\)~\cite{jahani2022berrut}. For a fair comparison, all other hyperparameters, including the smoothing parameters of the encoder and decoder splines, are kept identical across both settings. Each experiment is repeated \(10\) times, and we report the average and \(95\%\) confidence interval, as depicted in Fig.~\ref{fig:gcc_comparison}.

\paragraph{Evaluation Metrics}
We evaluate performance using mean squared error (MSE):
\(
    \mathrm{MSE} = \mathbb{E}_{\mathcal{X},\mathcal{F}} [ \frac{1}{K} \sum_{k=1}^K \|\func(\mathbf{x}_k) - \widehat{\func}(\mathbf{x}_k)\|_2^2 ],
\)
averaged over input batches \(\mathcal{X}\) and straggling sets \(\mathcal{F}\). For LeNet5 classification, we additionally measure relative accuracy, \(\mathrm{RelAcc} = \mathrm{Acc}(\widehat{\func}) / \mathrm{Acc}(\func)\), where \(\mathrm{Acc}(\cdot)\) denotes classification accuracy on the input batch.
\paragraph{Results}
Fig.~\ref{fig:gcc_comparison} presents the performance comparison of GCC Manifold and GCC. From left to right, the first two panels report the LeNet5 results with \((N,K)=(100,60)\). The proposed manifold-aware ordering consistently achieves lower MSE (Fig.~\ref{fig:lenet5_mse}) and higher relative accuracy (Fig.~\ref{fig:lenet5_relacc}) than standard GCC, with the improvement becoming more pronounced in the high-straggler regime. The third panel reports the MSE for the degree-\(8\) polynomial task with \((N,K)=(31,12)\) (Fig.~\ref{fig:poly_mse}), while the fourth panel provides a zoomed-in view for fewer than \(9\) stragglers (Fig.~\ref{fig:poly_mse_zoom}). The manifold-aware method exhibits more stable and better recovery behavior. While standard GCC shows greater variability and increasing recovery error as the number of stragglers grows, the proposed method maintains a recovery error close to zero over the tested straggler range.

% \section{Conclusion}
% \label{sec:conclusion}
% This paper introduced a manifold-aware encoding strategy for general coded computing. The key idea is that coded-computing schemes for high-dimensional computations, including modern machine learning workloads, should exploit not only redundancy across workers but also structure in the input data. Although the master node cannot freely replace the inputs with an arbitrary compressed representation, the low-dimensional geometry often present in high-dimensional data can still guide the construction of coded inputs.

% The proposed method constructs a distance-weighted graph on the input batch and orders the data through a short Hamiltonian path before fitting the GCC encoder. This encourages the encoder curve to follow the empirical geometry of the data while preserving the standard encoding, computation, and decoding stages of GCC. Experiments on LeNet5 inference and high-dimensional polynomial evaluation demonstrate that this geometry-aware modification consistently improves recovery performance under straggling. More broadly, these results suggest that input-data geometry is a valuable design resource for coded computation.

\section*{Acknowledgment}
This work is supported by the National Science Foundation under Grant CIF-2348638.

\bibliographystyle{IEEEtran}

\end{document}